\documentclass{article}

\usepackage{graphicx}
\usepackage[hidelinks]{hyperref}
\usepackage[a4paper, margin=2.5cm]{geometry}

\AtBeginDocument{%
  }

\begin{document}

\title{Looking for the Human in HRI Teaching:\\User-Centered Course Design\\for Tech-Savvy Students}

\author{Tobias Doernbach\\Human-Centered Robotics Lab\\Ostfalia University of Applied Sciences, Wolfenbuettel, Germany}
\date{}


\newcommand{\todo}[1]{\textbf{\textcolor{red}{TODO: #1}}}
\newcommand{\toolname}{\emph{\textsf{Node-(RED)}$\mathsf{\mathit{^2}}$}}

\maketitle

\begin{abstract}
Top-down, user-centered thinking is not typically a strength of all students, especially tech-savvy computer science-related ones. We propose Human-Robot Interaction (HRI) introductory courses as a highly suitable opportunity to foster these important skills since the HRI discipline includes a focus on humans as users.
Our HRI course therefore contains elements like scenario-based design of laboratory projects, discussing and merging ideas and other self-empowerment techniques. Participants describe, implement and present everyday scenarios using Pepper robots and our customized open-source visual programming tool. We observe that students obtain a good grasp of the taught topics and improve their user-centered thinking skills.
\end{abstract}



\section{Introduction}
Human-Robot Interaction includes humans in its name, however, tech-savvy computer science-related students sometimes think in a \emph{here is the solution, where is the problem} way. Not only with customer-focused development approaches like Scrum, engineers need to be able to think like and for their customer in order to deliver the right product.

Therefore, apart from teaching basic HRI topics like verbal and non-verbal interaction, spatial interaction and emotions, the \emph{Human-Robot Interaction} introductory course taught at Ostfalia University of Applied Sciences for the Computer Science Master of Science program uses Scenario-Based Design elements to achieve user-centering. We set the following learning goals:
\begin{itemize}
    \item being able to think top-down, from a user/customer perspective. This is typically not a main strength of rather tech-savvy students and such a type of course is a perfect opportunity to teach thinking user-first.
    \item finding a relevant problem and defining a solution for it, instead of simply implementing the path towards a solution of a given problem.
    \item discussing, merging and iterating on similar or different ideas in a group.
    \item fostering participants' understanding that they are empowered to take any decision about their self-defined scenarios and technical solutions and add or change anything they deem necessary, as long as they are able to explain their line of thought and why this is relevant to solve the initial problem.
    \item being able to evaluate approaches and solutions in a user-centered way, using user studies.
    \item using Generative AI as a text generation tool in a productive and reflected way.
\end{itemize}
How the technical solution exactly looks like is not a main learning of this course; we rather emphasize the route to the solution and try to engage critical thinking and self-empowerment, as explained in the chronological course breakdown in the following sections.

\section{Course Organization}
The course is based on three main components: \textbf{HRI contentual topics} delivered in a mixed lecture/\allowbreak discussions/\allowbreak group-work format, review and presentation of a \textbf{scientific publication}, and design and lab-stage implementation of a comprehensive \textbf{everyday HRI scenario} using a Pepper robot.
Time-wise, three-hour weekly sessions have been arranged so that theoretical and practical content can be shifted dynamically. This turned out to be a very efficient way of allowing enough time for in-class help with practical topics, but also to make sure that enough time could be devoted where theoretical parts required further explanation. In the second half of the course, full lab sessions are held every other week without theoretical contents  so that enough assistance could be provided and students were also aware that enough of their time should go into the work on their lab scenario.

\section{HRI Intro Contentual Topics}
The content storyline is based on the seminal book "Human-Robot Interaction -- An Introduction" by Bartneck et al.~\cite{bartneck2020human}. In past iterations of the course, we have taught these in lecture style with exercises and discussions in between. However, for the next iteration this summer semester, we plan to run the theoretical content in a flipped-classroom style, with requiring the respective reading to be done before the class and answering questions and providing discussions during class. The rationale is to encourage more reading instead of a pure consumative attitude which, unfortunately, has happened in the past, possibly due to the missing final written or oral exam. One entry point for these discussions are topically appropriate research publications as to be presented by students.

\section{Research Paper Presentations}
All students are required to select one recent peer-reviewed research paper from a provided list, review it and give a 10-minute about its contents with their fellow students as the target group (some examples: \cite{ostrowskiDesignResearchHRI2020,kirschnervoluntaryMotionHumanRobot2021,wieseEmbodiedSocialRobots2018, erelExcludedRobotsCan2021,caoInvestigatingRoleMultimodal2023,andreassonAffectiveTouchHuman2018}) . They confirm their selection by submitting their choice into a time planning tool where also the presentation dates are fixed depending on the contentual fit.

This part of the class has the goal to make the students aware of how to read a research paper, extract the essence and process it for a different audience.
An additional constraint of the presentation is to not use any words or characters on slides except for the title slide, one headline and anything embedded in images cited from the papers. This encounters the classical "wall-of-text" approach inexperienced students often take to prepare presentations.
Participants showed great creativity with this constraint; some of them showed videos or self-made animations, some used loopholes of live-writing onto the slide or a flipchart (which was permitted) etc.

\section{Lab Scenario}
\begin{figure}
    \centering
    \includegraphics[width=0.4\linewidth]{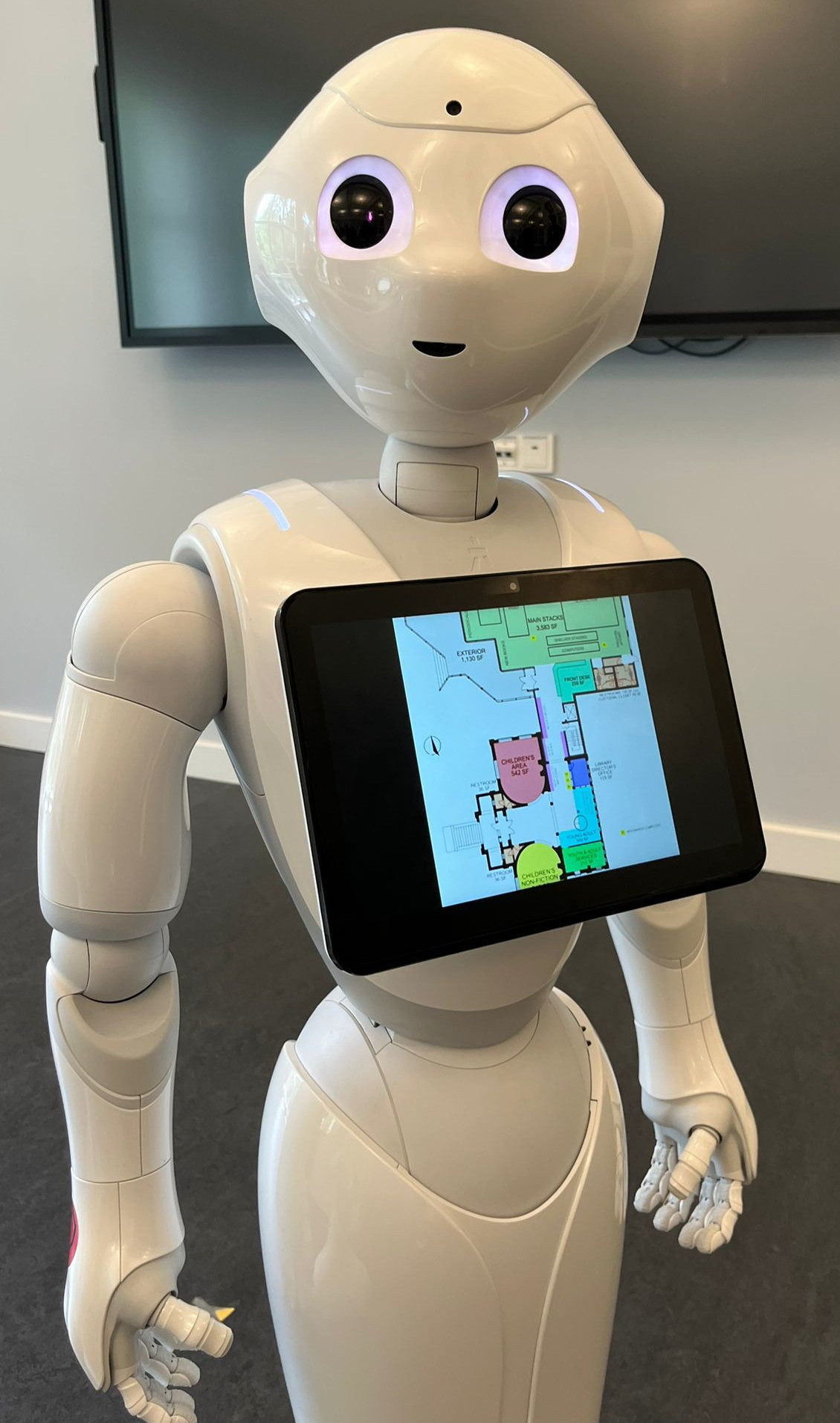}
    \caption{Pepper as a library assistant showing a map of the building}
    \label{fig:pepper-librarian}
\end{figure}

The Lab Scenario is a per-group implementation of a realistic everyday task in a domain defined by the group members, e.g.\ in a library as seen in Fig.~\ref{fig:pepper-librarian}. A live demonstration of a restaurant order scenario using a Pepper robot is presented in the first class session as a guideline how such a scenario could look like. Using these as an inspiration, participants are asked to (individually at first) select and describe a scenario of appropriate size and complexity for the given timeframe, and to be described in a proposal.

\subsection{Lab Scenario Proposal}
We employ Scenario-Based Design as a subtopic of User-Centered Design as our main design instrument to encourage top-down thinking and the other learning goals outlined in the introduction.
In order to efficiently begin with the definition of scenarios given the course's time constraints and requirements of technical work, students are not explicitly required to use a strict persona-oriented approach, but encouraged to do so. The first task, within the first two weeks of the course, is to write an initial proposal for a lab scenario to be implemented, which is later improved in group-work phase.

\subsubsection{Proposal First Stage: Individual Work}
After the first two weeks of the course, students have to submit their individual first-stage proposal which then is taken as an input for the directly following Lab Group Selection.

The main things students are supposed to describe are:
\begin{itemize}
    \item usefulness - how would anyone use this and why?
    \item which problem does the idea solve?
    \item how does the solution (roughly) work -- brief description of the architecture/components.
\end{itemize}
Because having taken a robotics class is not a mandatory requirement and the architecture may change during the course of the lab, participants are not required to show deep understanding about the architecture yet.

Instead, they are supposed to already think and include draft ideas in each of the categories of \emph{verbal interaction}, \emph{non-verbal interaction}, \emph{spatial interaction}, and \emph{emotion}.
Since these contentual topics are only covered during the class, an educated guess is sufficient for how to concretely include these categories.

\subsubsection{Lab Group Selection}\label{group_selection}
One common issue of group-based lab work usually is the group selection mechanism. "Classical" mechanisms as used in this class previous semesters, apart from completely random group assignment, are a) free choice of group partners (typical main issue: better and worse students form groups respectively, so that no mutual support within the group can happen) and b) skill-based selection after filling in a self-assessment skill questionnaire (typical main issue: students complain that they end up with "random" people they don't want to work with for whatever reason).

What we attempted successfully in a different class in winter semester 2023 is the following group selection mechanism:
After submitting their individual scenario proposals and we gave some initial feedback, the next class session is designed as a "proposal fair" where, in several sessions with around 8-10 students each, every students presents one "poster" (slide) about their proposal, all other students are encouraged to walk around, talk to the presenters and ask questions. The instructor does the same until everyone knows all other options, then the next session begins with the next 8-10 students and this is repeated until all students presented their proposal.

This way, everyone gets to learn about most of the other approaches. Additionally, all participants are to upload their "poster" in the course so that everyone can review them. Until the next session, students are supposed to form lab groups and submit their choice in the group selection tool. In case of ties, undecided students or other difficulties, this is moderated by the instructor at the beginning of the next class. In addition, the instructor is able to ask any students to leave the class who have not contributed so far or did not show up to the fair, being able to continue with active students only.

Compared to the previous semesters which employed the described "classical" selection mechanisms, this generated mainly positive feedback by students. Additionally, we were able to observe non-re\-pre\-sen\-ta\-tive\-ly that some students that we knew as rather weak were profiting a lot from their peer. Almost no group work issues were reported or observed by me, definitely less than in previous semesters.

\subsubsection{Proposal Second Stage: Group Work}
Once the lab groups have converged, they are asked to merge and revise their first-stage proposals using the comments given by fellow participants and the instructor. This proposal eventually serves as a guideline for the lab scenario to be implemented, sets the baseline for the group's aspiration and the final solution will be checked versus this document to confirm whether the self-defined claims have been met.

\subsubsection{User Study}
Within the second-stage proposal, a user study has to be described that could be executed\footnote{Running the user study is not part of the course for time reasons, however, good students are encouraged to continue their work in their final thesis and run the study using this description as a basis.} to evaluate the implemented scenario. The learning objective is to enable students to prepare experiments with users in the loop which usually involves a user study. This way of evaluating technical and scientific developments is usually not well-known to students in a technical major.
Hoffman and Zhao~\cite{hoffmanPrimerConductingExperiments2020} give an excellent overview which steps are required to design, conduct and evaluate the results of a comprehensive HRI user study. Therefore, students are required to read this paper and describe the necessary steps to run a scientifically sound user study on their lab scenario.

\subsection{Lab Scenario Implementation}
For the implementation of the scenario, two Pepper robots have been provided. Concurrent robot usage is moderated by an online booking tool\footnote{\url{https://booking.robotics-i.ostfalia.de/?locale=en-US}}. Students can freely access the lab using their university ID during daytime hours after a mandatory safety briefing. They are supposed and motivated to work outside of official class hours as well.

For this course, participants were supposed to use the open-source visual programming tool \toolname \cite{weikeEnablingUntrainedUsers2024} developed within the HCR Lab which is freely available on\\\url{https://github.com/robotics-empowerment-designer}. This tool is based on the \emph{Node-RED} visual programming Internet of Things (IoT) tool, has been adapted to be used with Pepper and is currently being extended for use with multiple different robots and IoT devices. Please see \cite{weikeEnablingUntrainedUsers2024} for further explanations and the rationale of developing this tool.

\subsection{Lab Scenario Demo (Open Lab)}
In the end of the semester, participants demonstrate their solutions in groups. They are also asked to record a video of their solution as a backup in case of unforeseen technical issues.
In summer semester 2024, for the first time we plan to arrange all lab demos as an Open Lab day where all robotics-related courses demonstrate their lab projects and the whole department is invited. This has been hosted by other universities before in a similar way and creates a great opportunity to share ideas, to network and to interest other students in the topic and lab.

\subsection{Lab Scenario Review}
From the 2023 iteration of the course, we added Generative AI (specifically: Large Language Models/LLMs) as course contents, as a discussion as well as a final individual assignment. The idea is to embrace tools like ChatGPT as a tool for productive use while reflecting about how and for what to use it.
Therefore, the assignment includes re-creating the lab scenario description using an LLM, listing all prompts and addressing the questions listing in \ref{questions}. Instructions included to write 3000-4000 words while a) using at least three scientific sources to support the posted claims and b) not emply an LLM for this second part of the assignment
Surely, this assignment needs to be adapted to current developments when LLMs gain more and more traction.

\section{Course Grading and Difficulty Settings}
\begin{figure}
    \centering
    \includegraphics[width=\linewidth]{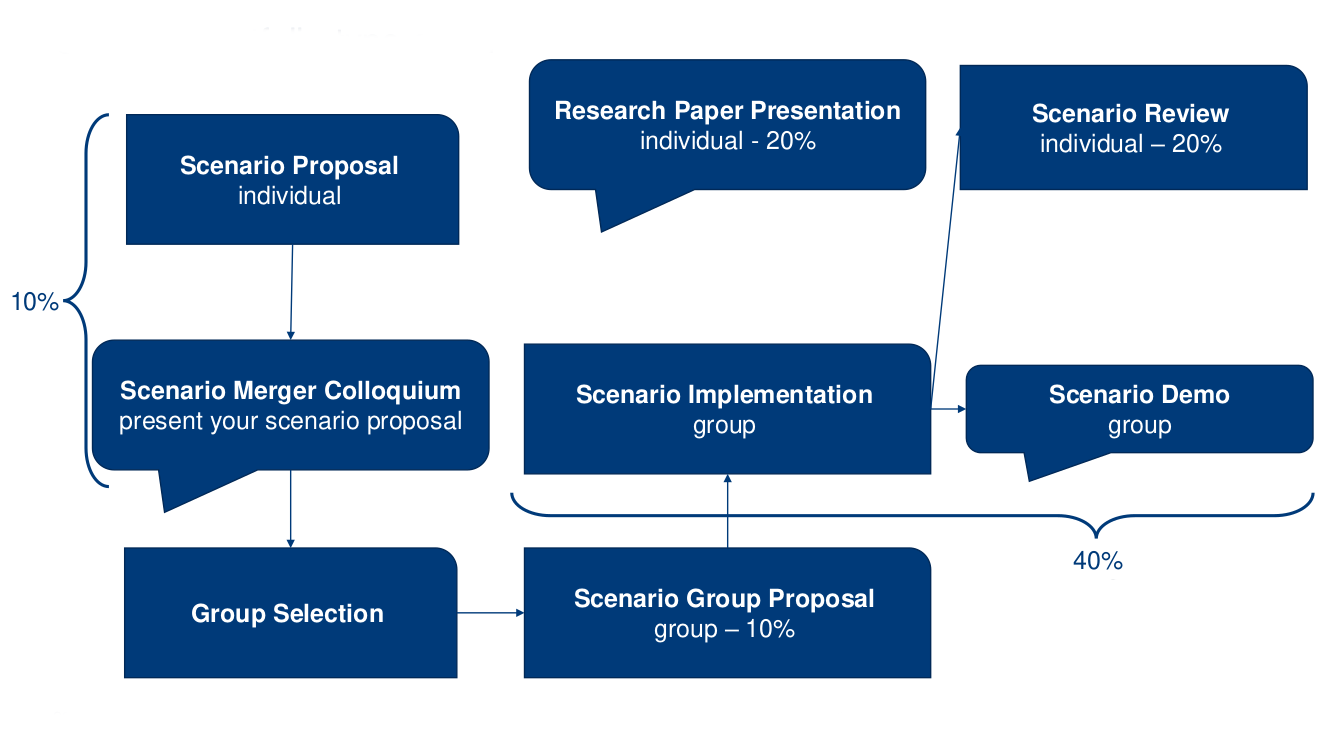}
    \caption{Grading scheme as presented to participants}
    \label{fig:grading}
\end{figure}
Because Human-Robot Interaction is an advanced topic where significant effort goes into lab work and previous technical knowledge may be diverse, group work is a main feature to achieve significant results. However, group-only work generates a grading problem where individuals may hide behind their peers which has been observed in previous iterations of the course. Hence, besides careful consideration of the lab group selection mechanism (see Section~\ref{group_selection}), individual grade parts bring back some focus on the individual.

Therefore, we decided to split the course grading into an individual and a group part which are both worth 50\%, however, 51\% are needed to pass with the lowest passing grade. This way, no student can "slip through" with excellent group members that achieve 100\% on the group part while not submitting anything individually. The artifacts and actions that contribute to the grade are shown in Fig.~\ref{fig:grading}.

Although this course has been designed and is currently in operation for graduate students, it can easily be scaled down to undergraduate level by asking for less in-depth understanding in the respective artifacts to submit during the course and the oral presentations.

One degree of freedom for the difficulty of the course is the level of the provided publications, e.g. \cite{ostrowskiDesignResearchHRI2020,kirschnervoluntaryMotionHumanRobot2021,wieseEmbodiedSocialRobots2018} on undergraduate level or \cite{erelExcludedRobotsCan2021,caoInvestigatingRoleMultimodal2023,andreassonAffectiveTouchHuman2018} on graduate level. This list of course can freely be adapted to the entry and/or desired learning level of the students and can be extended almost arbitrarily long for a rising number of students (limited only by the amount of adequate-level publications).
However, the length of the publications as well as the student-perceived difficulty optimally do not diverge too far in order to keep the difficulty of understanding the contents (and therefore delivering a good presentation) on comparable level.

With respect to the scenario difficulty, the instructor can easily shrink down or scale up the scenarios after submission of the first- and second-stage proposals. This is essential to fit the desired difficulty because, since many non-technical skills are required as well as technical topics which students may not have been in touch with, they often may not fit a reasonable scenario for the given time frame. This part of the course, in our experience, is the main job of the instructor to put emphasis on because overall success of the course strongly depends on their estimate of the scenario difficulty and time requirements.

\section{Results}
\begin{table}[t]
    \centering
    \begin{tabular}{r|l}
         \#groups & domain \\
         \hline
         7 & restaurant/bar waiter \\
         3 & hospital reception assistant \\
         2 & library assistant \\
         2 & hotel receptionist \\
         1 & HCR lab tour \\
         1 & dance instructor \\
         1 & movie theater service employee \\
    \end{tabular}
    \caption{Lab scenario domains of two course iterations as selected by participants}
    \label{table:domains}
    \vspace{-5mm}
\end{table}
In our experience, not all topics taught in class are covered in the lab scenario implementation in the same depth because of their difficulty and time requirements. It seems especially hard to embed emotions in a way that includes closed-loop display of emotions and verification of their usefulness. It can be said that not too-much in-depth solutions could be observed during the previous runs of this course, mainly because technical difficulties (apart from the usual issues like suboptimal time management) prohibit the majority of students to deliver something really sophisticated with respect to advanced topics. However, a basic understanding of all topics could be observed with the majority of participants, and of the relatively simpler topics of verbal interaction and accompanying gestures, students had a better grasp.

With respect to implemented scenarios, in the past two iterations of the course, student groups selected the domains in Table~\ref{table:domains}. We are going to encourage more non-obvious domains like dance instructors in the future to achieve even more creative thinking with the participants.

\section{Conclusion}
Since top-down thinking is not typically a strength of all students, an HRI course is the perfect opportunity to foster this. The topics and scenarios such a course deals with require a user-centered approach anyway, by the definition of the title of the discipline. Therefore, we encourage the community to embrace this opportunity to realize the full potential of their human-centered robotics teaching.


\bibliographystyle{IEEEtran}
\bibliography{bibliography}

\appendix

\section{Lab Scenario Report Questions}\label{questions}
\begin{itemize}
    \item How many tries did you need to get a useful result?
    \item Which strategy did you follow to get a satisfactory result (i.e. why did you use your prompts in this way?)
    \item What else could the LLM be used with regard to your lab scenario?
    \item Which attack/forgery probabilities do you see by using the LLM in your lab scenario?
    \item Which prompt has been surprisingly successful/unsuccessful?
    \item How would you rate the quality of the answers?
    \item How would you rate the work with the LLM? What did you like/dislike?
    \item What is the most important aspect when creating a prompt in your opinion?
    \item What are the limitations of the LLM in your opinion? Name a few areas where it is not recommended to be used.
\end{itemize}

\end{document}